\documentclass[ a4paper, 10 pt, twocolumn]{IEEEtran}                                                          

\usepackage[export]{adjustbox}
\usepackage{tabularx}
\usepackage{subfigure}
\usepackage{amsmath,amssymb}
\usepackage{latexsym}
\usepackage{psfrag}
\usepackage{epsfig}
\usepackage{graphicx}
\usepackage{pifont}
\usepackage{array}
\usepackage{collref}
\usepackage{enumerate}
\usepackage{cite}
\usepackage{amsmath}
\usepackage{amscd}
\usepackage{amssymb}
\usepackage{latexsym}
\usepackage{array}
\usepackage{tabularx}
\usepackage[ruled]{algorithm}
\usepackage{algorithmic}
\usepackage{epsfig,subfigure}
\usepackage{changepage}
\usepackage{stfloats}
\usepackage{multirow}
\usepackage{multirow}
\usepackage{bigstrut}
\usepackage{array}
\usepackage{graphicx}
\usepackage{fancyhdr}

\usepackage{algorithm}
\usepackage{color}

\allowdisplaybreaks
\allowdisplaybreaks[1]

\newtheorem{theorem}{Theorem}

\title{\bf A Non-linear Reweighted Total Variation Image Reconstruction Algorithm for Electrical Capacitance Tomography}

\author{Kezhi~Li, Daniel J. Holland %
\thanks{ Imperial College London, UK; Dept. of Chemical and Process Engineering, University of Canterbury, New Zealand; Magnetic Resonance Research Centre (MRRC), University of Cambridge, UK. Email: kezhi.li@imperial.ac.uk; daniel.holland@canterbury.ac.nz}}

\begin{document}
\maketitle

\begin{abstract}
A new iterative image reconstruction algorithm for electrical capacitance tomography (ECT) is proposed that is based on iterative soft thresholding of a total variation penalty and adaptive reweighted compressive sensing. This algorithm encourages sharp changes in the ECT image and overcomes the disadvantage of the $l_1$ minimization by equipping the total variation with an adaptive weighting depending on the reconstructed image. Moreover, the non-linear effect is also partially reduced due to the adoption of an updated sensitivity matrix. Simulation results show that the proposed algorithm recovers ECT images more precisely than existing state-of-the-art algorithms and therefore is suitable for the imaging of multiphase systems in industrial or medical applications.

\end{abstract}


\section{Introduction}

Electrical capacitance tomography (ECT) is an attractive method for imaging multiphase flows, as it is noninvasive, fast, safe and low cost \cite{Yang-ECT_review,Marashdeh-ECT_review}.
A typical ECT system consists of three main parts: a multi-electrode sensor, an acquisition hardware and a computer for hardware control and image processing. Specifically, the multi-electrode hardware in ECT typically has $n$ electrodes surrounding the wall of the process vessel. The number of independent capacitance measurements in such a configuration is $N=1/2\cdot n(n-1)$ due to the independent number of sensor pairs with $n$ electrodes. The final objective is to recover the cross section or even 3D images of the permittivity distribution by using these measurements to solve an inverse problem. However, the inverse problem is underdetermined, since the number of measurements is far fewer than the number of pixels in the reconstructed image \cite{Phua-WeakInv}. Furthermore, the governing equations to be considered are non-linear \cite{Marashdeh-Nonlinear, Haddadi-ANewAlgForImg}.  Various reconstruction algorithms have been developed to cope with these difficulties. Direct, or single step, algorithms include the classic linear back projection (LBP), approaches based on Singular Value Decomposition (SVD), and Tikhonov regularization \cite{Yang-ImageRec}. Indirect, or iterative, algorithms include Landweber iterations (LI) \cite{Yang-AnImageRec} and iterative Tikhonov methods \cite{Peng-UsingReg,Wang-AHighSpeed}. These algorithms all inherently assume a smooth permittivity distribution within the sample, but for many systems this assumption is poor. In recent years, concepts from compressive sensing (CS) theory\cite{donoho-cs,Candes_sparsity_incoherence} have been shown to permit the reconstruction of sharp changes in permittivity \cite{Chandrasekera-TotalVar,Ye-ImageRecForECT,Song-TVForEIT}. CS cannot be applied strictly to ECT image reconstruction due to the non-linear nature and because the sensitivity matrix does not satisfy the restricted isometry property (RIP). However, several researchers have extended the ideas of CS to non-linear systems \cite{Blumensath-NonLin, Ohlsson-NonLin, Xu-NonLin, Ehler-NonLin}.

In this paper we adapt ideas from CS to propose a comprehensive Reweighted Total Variation Iterative Shrinkage Thresholding (TV-IST) Algorithm for non-linear ECT image reconstruction. After explaining the ECT physical model, we modify the conventional TV-IST to develop the TV-IST for ECT and its fast version with auxiliaries. We introduce adaptive weights \cite{Candes-Enhancing} to approximate the $l_0$-norm solution closely. Finally, we combine this reweighting approach with a method to minimize the non-linearity of the reconstruction by updating the sensitivity matrices within TV-IST. The algorithm has been examined using simulated measurements of phantoms to show its superiority compared with other existing algorithms.

\section{Fundamentals of Electrical Capacitance Tomography}
In ECT, the permittivity distribution inside a pipe or vessel of interest, corresponding to the material distribution, is calculated from measured capacitances between all pairs of sensors located around the pipe's periphery.
The total electric flux over all the electrodes surfaces is equal to zero, hence the potential and permittivity are obtained from a form of Poisson's equation:
\begin{equation}
\nabla \cdot \left[ \epsilon(\mathbf{r})\nabla \phi(\mathbf{r})\right] = 0,
\end{equation}
where $\epsilon(\mathbf{r})$ is the spatial permittivity distribution, and $\phi(\mathbf{r})$ the electric potential distribution. The boundary conditions are $\phi=V_c$ for the excited electrode and $\phi=0$ for other electrodes.

For the two-dimensional case $\mathbf{r}=(x,y)$, the relationship between the capacitance and permittivity distribution can be expressed by the following equation:
\begin{equation}\label{eq:C}
C=\frac{Q}{V_c} = -\frac{1}{V_c} \oint_S \epsilon(x,y) \nabla \phi(x,y)\text{d}s,
\end{equation}
where $Q$ is the total charge, $S$ denotes the closed line of the electrical field, $\epsilon(x,y)$ is the permittivity distribution in the sensing field, and $V_c$ is the potential difference between two electrodes forming the capacitance.

In (\ref{eq:C}) ${\phi}(x,y)$ is also a function of $\epsilon$. Therefore the capacitance between electrode combinations can be considered as a function of permittivity distribution $\epsilon(x,y)$:
\begin{equation}
C= f(\epsilon),
\end{equation}
where $f$ is a non-linear function, and elements of $C$ are the non-redundant capacitance values obtained from the electrode pairs $\left[ C_{1,2}, C_{1,3}, \cdots, C_{1,n}, C_{2,3}\cdots C_{N-1,N} \right]$. If we descretise the permittivity and express it as a vector, we can estimate the changes in the capacitance values from a Taylor's series expansion:
\begin{equation}\label{eq: triangleC}
\triangle C = \frac{\text{d} f}{\text{d} \epsilon}(\triangle \epsilon) + O ((\triangle \epsilon)^2),
\end{equation}
where $\frac{\text{d} f}{\text{d} \epsilon}$ is the sensitivity of the capacitance with respect to changes in the permittivity distribution, and $O ((\triangle \epsilon)^2)$ represents the higher order terms of $(\triangle \epsilon)^2$. Because $\triangle \epsilon$ is usually small, the high order terms are often neglected. Then Eq. (\ref{eq: triangleC}) can be linearized in a matrix form:
\begin{equation}\label{eq:linearC}
\triangle \mathbf{C} = \mathbf{J} \triangle \boldsymbol{\epsilon},
\end{equation}
where $\triangle \mathbf{C} \in \mathbb{R}^M$, $\mathbf{J}\in \mathbb{R}^{M \times N}$ is a Jacobian/sensitivity matrix denoting the sensitivity distribution for each electrode pair, and $\triangle \mathbf{\epsilon} \in \mathbb{R}^{N}, N \gg M$. As a result, the non-linear forward problem has been reformulated to a linear approximation. Generally in ECT, Eq. (\ref{eq:linearC}) is written in a normalized form
\begin{equation}\label{eq:lambda}
\boldsymbol{\lambda} = \mathbf{S} \mathbf{x},
\end{equation}
where $\boldsymbol{\lambda} \in \mathbb{R}^M$ is the normalized capacitance vector, $\mathbf{S}\in \mathbb{R}^{M \times N}$ is the Jacobian matrix of the normalized capacitance with respect to the normalized permittivities, which gives a sensitivity map for each electrode pair, and $\mathbf{x} \in \mathbb{R}^{N}, N \gg M$ is the normalized permittivity vector, which can be visualized by the colour density of the image pixels. The conventional optimization problem of ECT becomes
\begin{equation}
\mathbf{x} = \arg \min_{\mathbf{x}}{||\boldsymbol{\lambda} - \mathbf{S} \mathbf{x}||^2}.
\end{equation}
Because there are $n$ electrode pairs, $M$ should be $1/2\cdot n(n-1)$. The objective of the reconstruction algorithm of ECT is to recover $\epsilon(x,y)$ from measured capacitance vector $C$. While in the discrete linear model, it is to estimate $\mathbf{x}$ given $\boldsymbol{\lambda}$, and $\mathbf{S}$ is seen as a constant matrix determined in advance for simplicity.


There are several difficulties with the reconstruction problem. Firstly, (\ref{eq:lambda}) is under-determined so the solution is not unique, and it is very sensitive to disturbances of $\lambda$. Secondly, owing to the non-linearity in Eq. (\ref{eq:C}), $\mathbf{S}$ is not constant but varies for different permittivity distributions.

In this paper, we propose a non-linear reweighted total variation image reconstruction algorithm to overcome these difficulties.
\section{Iterative Shrinkage Thresholding Algorithm Meets Total Variation}

\subsection{Iterative Shrinkage Thresholding Algorithm}
To recover the permittivity distribution image $ \mathbf{x}$, many reconstruction algorithms for ECT have been developed. Generally the reconstruction algorithms can be categorized in two groups: direct algorithms and iterative algorithms. Among them, Landwater Iteration and Steepest Descent Method (LWSDM) is considered as one of the best algorithms with good efficiency. It minimizes the cost function $\frac{1}{2} ||\mathbf{Sx}- \mathbf{\lambda}||^2_2$, e.g. to minimize
\begin{equation}
\begin{split}
f(\mathbf{x}) &= \frac{1}{2} (\mathbf{Sx}-\boldsymbol{\lambda})^T(\mathbf{Sx}-\boldsymbol{\lambda}) \\
&=\frac{1}{2}(\mathbf{x}^T\mathbf{S}^T\mathbf{Sx}-2\mathbf{x}^T\mathbf{S}^T\boldsymbol{\lambda}+\boldsymbol{\lambda}^T\boldsymbol{\lambda}). \end{split}
\end{equation}
The gradient of $f(\mathbf{x})$ is
\begin{equation}
\nabla f(\mathbf{x})=\mathbf{S}^T\mathbf{S}\mathbf{x} -\mathbf{S}^T\boldsymbol{\lambda}= \mathbf{S}^T (\mathbf{Sx}-\boldsymbol{\lambda}).
\end{equation}
We iteratively update the image $\mathbf{x}$ in the direction that $f(\mathbf{x})$ decreases most quickly. Therefore the new image will be
\begin{equation}
\mathbf{x}_{k+1}= \mathbf{x}_k - \alpha_k \nabla f(\mathbf{x}_k)=  \mathbf{x}_k - \alpha_k\mathbf{S}^T (\mathbf{Sx}_{k}-\boldsymbol{\lambda}),
\end{equation}
where $\alpha_k$ is a positive value determining the step size.

In fact, LWSDM can be derived from the Iterative Shrinkage Thresholding Algorithm (ISTA) as a special case with ECT constraints. Here we introduce and explain the general model of ISTA on which our algorithm also is based.

The ISTA is to solve a class of optimization problems with convex differentiable cost functions and convex regularization.
\begin{theorem}\label{ther:ISTA} \cite{Beck-AFastIterative} Consider the general formulation:
\begin{equation}\label{eq:xx}
\mathbf{x} = \arg \min_{\mathbf{x}}{ \left\{F(\mathbf{x}) \equiv f_1(\mathbf{x}) + f_2(\mathbf{x}) \right\}}, \mathbf{x} \in \mathbb{R}^N \ \ \ \ \ \ \ \ \ \ \ ( P_0)
\end{equation}
and the following assumptions are satisfied:
\begin{itemize}
  \item  $f_1$: a smooth convex function which is also continuously differentiable with Lipschitz continuous gradient $L(f_1)$:
  \begin{equation}
  ||\nabla f_1(\mathbf{x}) - f_1(\mathbf{y})  || \leq L(f_1)||\mathbf{x} - \mathbf{y}|| \ \ \ \ \text{for every\ } \mathbf{x},\mathbf{y} \in \mathbb{R}^N,
  \end{equation}
  where $L(f_1)>0$ is the Lipschitz constant of $\nabla f_1$.
  \item  $f_2$: a continuous convex function mapping $\mathbb{R}^N \rightarrow \mathbb{R}$
  \item  Problem $( P_0)$ is solvable.
  \end{itemize}
  Then basic ISTA converges to its true solution by running iteration $\mathbf{x}_k = p_L(\mathbf{x}_{k-1}), k = \{1,2, \cdots\}$, where its iteration:
  \begin{equation}\label{eq:p_L}
  p_L(\mathbf{y}) = \arg \min_{\mathbf{x}} \left\{ f_2(\mathbf{x}) + L/2 ||\mathbf{x} -\left( \mathbf{y} - 1/L \nabla f_1(\mathbf{y}) \right) ||_2^2 \right\}
  \end{equation}
\end{theorem}

For example, LWSDM is actually a special instance of problem $(P_0)$ by substituting  $f_1 := \frac{1}{2}||\boldsymbol{\lambda} - \mathbf{S} \mathbf{x}||^2$ and $f_2 := 0$ as a smooth quadratic minimization problem with the Lipschitz constant of the gradient $\nabla f_1$ being $L(f_1) = 2 \lambda_{max}(\mathbf{A}^T \mathbf{A})$. Then according to (\ref{eq:p_L}) we have
\begin{equation}\label{eq:x_k+1_f1}
\begin{split}
\mathbf{x}_{k+1} &= \arg \min_{\mathbf{x}} \left\{ L/2 ||\mathbf{x} -\left( \mathbf{x}_{k} - 1/L \nabla f_1(\mathbf{x}_{k}) \right) ||_2^2 \right\} \\
&= \mathbf{x}_k - \alpha \nabla f_1(\mathbf{x}_k)=  \mathbf{x}_k - \alpha\mathbf{S}^T (\mathbf{Sx}_{k}-\boldsymbol{\lambda}),
\end{split}
\end{equation}
which is equivalent to the LWSDM, where $\alpha = 1/L$. Theorem \ref{ther:ISTA} provides the theoretical convergence for algorithms.

\subsection{Total Variation Minimization}
Total variation (TV) norm of the image has been used widely to penalize the cost function \cite{Rudin-NonlinearTV}. It also has been verified that the TV norm can be utilized to address the under-determined image reconstruction and reproduce ECT  \cite{Soleimani-NonlinearImage, Chandrasekera-TotalVar} or other tomography images \cite{Liu-AdaptiveWeightedTV} with sharp transitions in intensity. Therefore, unlike conventional techniques for iterative reconstruction, we assume that there are sharp changes in intensity that can be sparsely represented by their spatial gradients. In this case the cost function is to minimize the least squares error and the sparsity of intensity changes:
\begin{equation}\label{eq:TV1}
\mathbf{x} = \arg \min_{\mathbf{x}}{ ||\boldsymbol{\lambda} - \mathbf{S} \mathbf{x}||^2 + \alpha||\mathbf{x}||_{TV}},
\end{equation}
where $||\mathbf{x}||_{TV}$ is the discrete isotropic TV of the two dimensional $\mathbf{X}= \text{reshape}(\mathbf{x},n_1,n_2) \in \mathbb{R}^{n_1 \times n_2}$ defined by \cite{Chambolle-AnAlgorithmFor}:
\begin{equation}
\begin{split}
||\mathbf{x}||_{TV} = &\sum_{i=1}^{n_2-1}\sum_{j=1}^{n_1-1}\sqrt{(\mathbf{X}_{i,j}-\mathbf{X}_{i+1,j})^2+(\mathbf{X}_{i,j}-\mathbf{X}_{i,j+1})^2}
\end{split}
\end{equation}
with the boundary conditions $\mathbf{X}_{n_2+1, j} - \mathbf{X}_{n_2,j} = 0, \forall j$ and $\mathbf{X}_{i, n_1+1} - \mathbf{X}_{i,n_1} = 0, \forall i$. (\ref{eq:TV1}) belongs to linear inverse problems with nonquadratic regularizers. Nonquadratic regularizers include wavelet representations \cite{Guerquin-AFastWavelet}, sparse regression \cite{Donoho-OptimallySparse} and total variation, etc. These problems can be solved by a signal processing technique called compressive sensing (CS) in the literature \cite{donoho-cs, Candes_sparsity_incoherence}. ISTA is very convenient to solve CS problems with $l_1$ norm regularization. The non-linear shrinkage operation, or so called soft thresholding, is
\begin{equation}\label{eq:shrinkage}
\begin{split}
\mathcal{T}_{\alpha} =  \left( |\mathbf{b}| - \min(\alpha, |\mathbf{b}|) \right) \cdot \text{sgn}(\mathbf{b}).
\end{split}
\end{equation}
For instance, the ISTA and its derivative versions along with the shrinkage operation have been verified to solve wavelet-based reconstruction for magnetic resonance imaging (MRI) efficiently \cite{Guerquin-AFastWavelet}. In the next section we will explain how to implement ISTA to ECT image reconstruction using TV regularization and prove its effectiveness.

\section{TV-IST for ECT}
In this section we will present an iterative reconstruction technique for ECT. As in IST, the iterative soft thresholding is applied to penalize the total variation of the ECT image. Some of the contents have been introduced in our conference paper \cite{Chandrasekera-TotalVar}. Here we provide the full theoretical analysis of this algorithm and its convergence rate.

In the ECT model, the permittivity distribution inside the pipe can usually be formulated as the 2D image/matrix $\mathbf{X}$. Set $\mathbf{x}$ is $\mathbf{X}$ expressed as a column vector. $\mathbf{X}_{i,j}$ denotes the pixel of the position $(i,j)$ in the imaging region. Its magnitude is proportional to the permittivity difference $\triangle \epsilon$ and is $0$ outside of the imaging region. We use $\mathbf{g}_1, \mathbf{g}_2$ to represent the gradients of the image respectively which correspond to the horizontal and vertical finite differences. In detail, the gradient transforms are used to calculate the gradients:
\begin{equation}\label{eq:g1g2}
\mathbf{g}_1 = \mathbf{G}_1 \mathbf{x}, \ \ \ \ \ \mathbf{g}_2 = \mathbf{G}_2 \mathbf{x},
\end{equation}
where $\mathbf{G}_1,\mathbf{G}_2$ are transform matrices. Each element of ${\mathbf{g}_1}^{(i)}, {\mathbf{g}_2}^{(i)}$ corresponds to the same $i$th element in $\mathbf{x}$. Likewise, given $\mathbf{g}_1,\mathbf{g}_2$, an inverse transform can be carried out by solving a least squares (LS) problem of
\begin{equation}\label{eq:x=g12}
\mathbf{x}(\mathbf{g}_1,\mathbf{g}_2) = \arg\min_{\mathbf{x}}\left\{||\mathbf{g}_1 - \mathbf{G}_1 \mathbf{x}||^2 + ||\mathbf{g}_2 - \mathbf{G}_2 \mathbf{x}||^2 \right\}.
\end{equation}
Using linear algebra we can obtain the standard LS solution:
\begin{equation}\label{eq:LS}
\begin{split}
\mathbf{x} &= \mathbf{L}^{-1}(\mathbf{G}_1^T \mathbf{g}_1 + \mathbf{G}_2^T \mathbf{g}_2), \\
& \text{where} \ \mathbf{L} = \mathbf{G}_1^T \mathbf{G}_1 + \mathbf{G}_2^T \mathbf{G}_2.
\end{split}
\end{equation}
$\mathbf{L}$ approximates the Laplacian operator for the image. It is an approximation of the Fourier transform version \cite{Michailovich-AnIter} but only considers the pixels within the imaging region.

\begin{figure*}[bt]
   \centering
   \begin{minipage}[t]{0.49\linewidth}
   \centering
  \includegraphics[width=6cm]{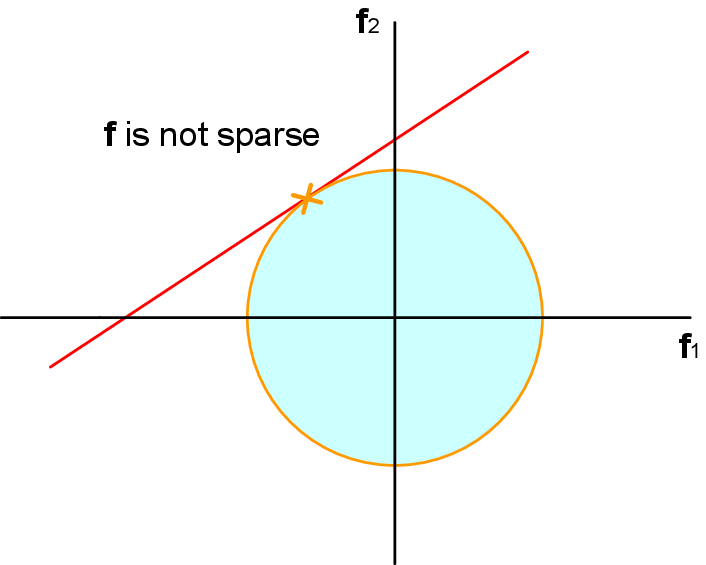} 
\centerline{(a)} 
   \end{minipage}
   \begin{minipage}[t]{0.49\linewidth}
   \centering
   \includegraphics[width=6cm]{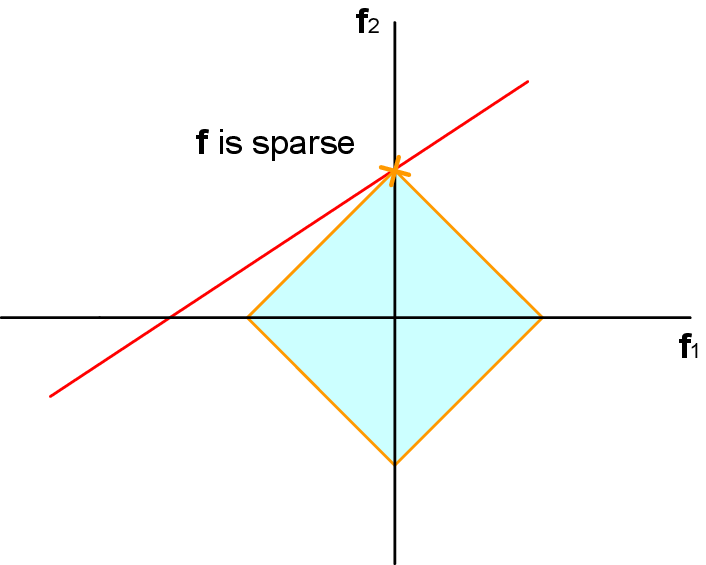} 
\centerline{(b)} 
   \end{minipage}
  \caption{An illustrative example indicating the advantage of $l_1$ minimization
over $l_2$ minimization in finding a sparse point in the line $\mathbf{y}= \mathbf{\Theta} \mathbf{f}'$.}\label{fig:L0L1}
\end{figure*}

To consider an isotropic form of TV, a single vector $\mathbf{g}$ is used to represent the gradient magnitude, where elements of $\mathbf{g}$ are given by $\mathbf{g}^{(i)} = \sqrt{{{\mathbf{g}_1}^{(i)}}^2 + {{\mathbf{g}_2}^{(i)}}^2}$. Then (\ref{eq:TV1}) can be reformulated to
\begin{equation}\label{eq:TVIST1}
\mathbf{x} = \arg \min_{\mathbf{x}}{ ||\boldsymbol{\lambda} - \mathbf{S}\mathbf{L}^{-1}(\mathbf{G}^T_1 \mathbf{g}_1 + \mathbf{G}_2^T \mathbf{g}_2)||^2 + \alpha||\mathbf{g}||_{1}},
\end{equation}
Equation (\ref{eq:TVIST1}) is different from the conventional $l_1$ minimization problem. However, we can still use the iterative update idea to pursue the solution. Instead of updating $\mathbf{x}_k$ in iterations, here we calculate $\mathbf{g}_1$ and $\mathbf{g}_2$ by updating it to their steepest descent. Following the second step in (\ref{eq:x_k+1_f1}), the residuals are calculated and projected to the $\mathbf{g}_1$ and $\mathbf{g}_2$ directions, respectively
\begin{equation}\label{eq:g12}
\begin{split}
\nabla f_1(\mathbf{g}_1)_1 &= \mathbf{G}_1 \mathbf{L}^{-1} \mathbf{S}^T (\mathbf{S} \mathbf{x}_k  - \boldsymbol{\lambda}), \\
\nabla f_1(\mathbf{g}_2)_2 &= \mathbf{G}_2 \mathbf{L}^{-1} \mathbf{S}^T (\mathbf{S} \mathbf{x}_k  - \boldsymbol{\lambda}),
\end{split}
\end{equation}
where $\{\mathbf{L}^{-1}\}^T = \mathbf{L}^{-1}$ due to the symmetry of matrix $\mathbf{L}$. As a result, according to the IST algorithm the gradients $\mathbf{g}_1,\mathbf{g}_2$ can be updated:
\begin{equation}\label{eq:hat_g12}
\begin{split}
\hat{\mathbf{g}_1}_{k+1} &= {\mathbf{g}_1}_{k} - \beta  \mathbf{G}_1 \mathbf{L}^{-1} \mathbf{S}^T (\mathbf{S} \mathbf{x}_k  - \boldsymbol{\lambda}), \\
\hat{\mathbf{g}_2}_{k+1} &= {\mathbf{g}_2}_{k} - \beta \mathbf{G}_2 \mathbf{L}^{-1} \mathbf{S}^T (\mathbf{S} \mathbf{x}_k  - \boldsymbol{\lambda}),
\end{split}
\end{equation}
where $\beta \leq 1/ \lambda_{max}(\{\mathbf{S}  \mathbf{L}^{-1}\mathbf{G}_i\}^T\{\mathbf{S}  \mathbf{L}^{-1}\mathbf{G}_i\})$ due to the requirements of Lipschitz continuous, and $\hat{\mathbf{g}}_{k+1}$ are the iterative gradient solution that can be derived from $\hat{\mathbf{g}_1}_{k+1}, \hat{\mathbf{g}_2}_{k+1}$ if we only consider the least squares error.

The next step is to optimize the $\mathbf{g}$ with
\begin{equation}
\mathbf{g}_{k+1} =  \arg \min_{\mathbf{g}}{||\mathbf{g}-\hat{\mathbf{g}}_{k+1}||^2 + \alpha' ||\mathbf{g}||_1 },
\end{equation}
where $\alpha'$ is equal to $\alpha$ multiplied by some constant. Similar to (\ref{eq:shrinkage}), a shrinkage operator can be used. While here the difference to the conventional shrinkage is that rather than set a soft thresholding on $\hat{\mathbf{g}}_1, \hat{\mathbf{g}}_2$ directly, we decrease the magnitudes of $\hat{\mathbf{g}}_1, \hat{\mathbf{g}}_2$ by making it proportional to the magnitudes of $\hat{\mathbf{g}}$ after a soft thresholding on $\hat{\mathbf{g}}$ to reduce the total variation. The magnitude vector $\hat{\mathbf{g}}$ can be calculated element-wise by
\begin{equation}\label{eq:g=g1g2}
\hat{\mathbf{g}}^{(i)} = \sqrt{\hat{\mathbf{g}_1}^{(i)2} + \hat{\mathbf{g}_2}^{(i)2}}.
\end{equation}
So for the ECT TV-IST, the soft thresholding process can be carried out
\begin{equation}\label{eq:soft-th-g12}
\begin{split}
{\mathbf{g}_1}^{(i)}_{k+1} &= \frac{\mathcal{T}_{\alpha'}(\hat{\mathbf{g}}^{(i)}_{k+1})}{\hat{\mathbf{g}}^{(i)}_{k+1}} \hat{\mathbf{g}_1}^{(i)}_{k+1}  \\
{\mathbf{g}_2}^{(i)}_{k+1} &= \frac{\mathcal{T}_{\alpha'}(\hat{\mathbf{g}}^{(i)}_{k+1})}{\hat{\mathbf{g}}^{(i)}_{k+1}} \hat{\mathbf{g}_2}^{(i)}_{k+1}
\end{split}
\end{equation}
where $\mathcal{T}_{\alpha'}$ is the shrinkage operator defined in (\ref{eq:shrinkage}), and this equation is calculated element-wise. By using this new soft thresholding process we are able to eliminate small variation and meanwhile reduce the large variation in $\hat{\mathbf{g}}_1, \hat{\mathbf{g}}_2$ directions. Finally, the new reconstructed image can be updated by (\ref{eq:LS}), and the new gradients of the image are updated by multiplying the transform matrices
\begin{equation}\label{eq:gi}
{\mathbf{g}_i}_{k+1} = \mathbf{G}_i \mathbf{x}, \ i \in{1,2},
\end{equation}
which returns to the beginning of the section and completes one iteration in the algorithm. Algorithm 1 sums up the Total Variation-Iterative Soft Thresholding algorithm. \\

\begin{adjustwidth}{-0.1cm}{}
\begin{tabular}{l}
\hline
\textbf{Algorithm 1}: TV-IST \\
\hline
\textbf{Input}: normalized sensitivity matrix $\mathbf{S}$, normalized \\ capacitance $\boldsymbol{\lambda}$, transform matrices $\mathbf{G}_1, \mathbf{G}_2$.  \\
\textbf{Set}: max loop $k_{\text{max}}$, shrinkage parameter $\beta$, $\alpha'$. \\
\textbf{Initialize}: $\mathbf{g}_1, \mathbf{g}_2, \mathbf{x}_0$ are zero vectors. \\
\textbf{Iteration}: for $k = 0,1, \cdots$, $k_\text{max}$ \ \ do \\
1. (\ref{eq:g12}) $\backslash \backslash$ calculate the steepest descent increment  \\
2. (\ref{eq:hat_g12}) $\backslash \backslash$ update $\mathbf{g}_1, \mathbf{g}_2$  \\
3. (\ref{eq:g=g1g2}) $\backslash \backslash$ update $\mathbf{g}$ element-wise \\
4. (\ref{eq:soft-th-g12}) $\backslash \backslash$ soft threshloding $\mathbf{g}_1, \mathbf{g}_2$ according to $\mathbf{g}$ \\
5. (\ref{eq:LS}) $\backslash \backslash$ update new image \\
6. (\ref{eq:gi}) $\backslash \backslash$ calculate new $\mathbf{g}_1, \mathbf{g}_2$ \\
\textbf{Output}: $\mathbf{x}$ or $\mathbf{X}$. \\
\hline\\
\end{tabular}
\end{adjustwidth}

In Algorithm 1, the iterative soft thresholding involves iterating with alternating the updates in (\ref{eq:hat_g12}), (\ref{eq:soft-th-g12}) and (\ref{eq:gi}). These procedures are performed for isotropic total variation to reduce image artifacts. Because the conservative property of the gradient vector field might be destroyed by the soft thresholding, (\ref{eq:gi}) is enforced at the end of each iteration \cite{Michailovich-AnIter}.

In addition, we also accelerate the convergence rate of the algorithm by adding the auxiliary vectors in iterations \cite{Beck-AFastIterative}. The auxiliary $\mathbf{h}_1, \mathbf{h}_2$ are updated based on $\mathbf{g}_1, \mathbf{g}_2$.
This technique is mature and has been used efficiently in various image processing areas \cite{Michailovich-AnIter, Wang-ANewAlter,Beck-FastGraBasedAlg}, called the Fast Iterative Shrinkage-Thresholding Algorithm (FISTA). It can be obtained from Algorithm 1 with several additional steps accordingly. \\

\begin{figure*}[ht]
   \centering
  \includegraphics[width=12cm]{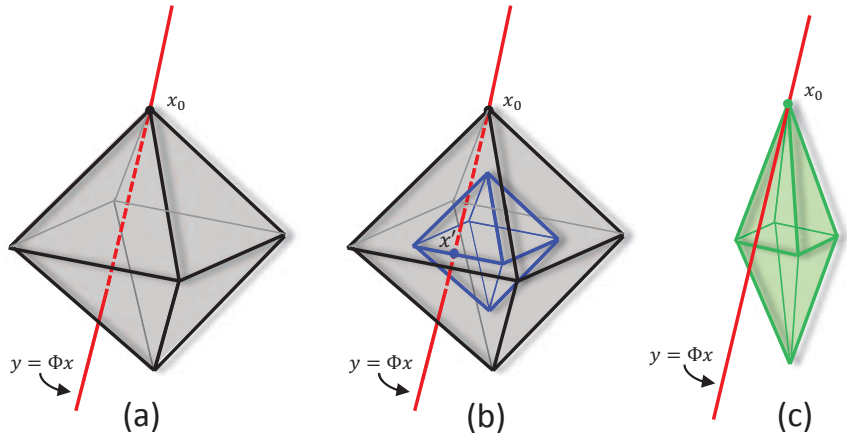} 
  \caption{An illustrative example indicating the advantage of weighted $l_1$ minimization
over conventional $l_1$ minimization in finding a sparse point in the line $\mathbf{y}= \mathbf{\Theta} \mathbf{x}'$.}\label{fig:wL1}
\end{figure*}

\begin{adjustwidth}{-0.1cm}{}
\begin{tabular}{l}
\hline
\textbf{Algorithm 2}: TV-FIST \\
\hline
\textbf{Input}: normalized sensitivity matrix $\mathbf{S}$, \\ normalized capacitance $\boldsymbol{\lambda}$, transform matrices $\mathbf{G}_1, \mathbf{G}_2$. \\
\textbf{Set}: max loop $k_{\text{max}}$, shrinkage parameter $\beta$, $\alpha'$. \\
\textbf{Initialize}: $\mathbf{g}_1, \mathbf{g}_2, \mathbf{h}_1, \mathbf{h}_2$ are zero vectors, $t_0 =1$. \\
\textbf{Iteration}: for $k = 0,1, \cdots$, $k_\text{max}$ \ \ do \\
1. ${\mathbf{x}}_{k+1} = \mathbf{L}^{-1}(\mathbf{G}_1^T \mathbf{h}_1 + \mathbf{G}_2^T \mathbf{h}_2)$, $\backslash \backslash$ update the image \\ according to auxiliary vectors \\
2. (\ref{eq:g12}) $\backslash \backslash$ calculate the steepest descent increment \\
3. $\hat{\mathbf{g}_1}_{k+1} = {\mathbf{h}_1}_{k} - \alpha' \nabla {f_1}_1 $, $\hat{\mathbf{g}_2}_{k+1} = {\mathbf{h}_2}_{k} - \alpha' \nabla {f_1}_2 $, \\
4. (\ref{eq:g=g1g2}), (\ref{eq:soft-th-g12}),(\ref{eq:LS}),(\ref{eq:gi})\\
5. ${t}_{k+1} = \frac{1+\sqrt{1+4t_{k}^2}}{2}$,  $ \backslash \backslash$  calculate $t$\\
6. ${\mathbf{h}_i}_{k+1} = {\mathbf{g}_i}{k+1} + \left( \frac{t_{k}-1}{t_{k+1}} \right)({\mathbf{g}_i}_{k+1}- {\mathbf{g}_i}_{k}), i \in \{1,2\}$, \\ $ \backslash \backslash$  calculate $\mathbf{h}_1, \mathbf{h}_2$ \\
\textbf{Output}: $\mathbf{x}$ or $\mathbf{X}$. \\
\hline\\
\end{tabular}
\end{adjustwidth}

Algorithm 2 for TV-FIST is an advanced version of TV-IST. It has been proved that FISTA can achieve a better rate of convergence of $\mathcal{O}(1/k^2)$ versus the IST algorithm's rate of $\mathcal{O}(1/k)$\cite{Beck-AFastIterative, Beck-FastGraBasedAlg}. Specifically, let $\{\mathbf{x}_n \}, \{\mathbf{y}_n \}$ be the sequence generated by Algorithm 2 with $\alpha$ satisfying the conditions in Theorem \ref{ther:ISTA}. Then for any $k \geq k' \in \mathbb{N}$, we have
\begin{equation}
\begin{split}
F(\mathbf{x}_k) - F(\mathbf{x}^*) &\leq \frac{c(\alpha)}{k-k'} ||\mathbf{x}_k-\mathbf{x}^*||^2,  \\
F(\mathbf{y}_k) - F(\mathbf{y}^*) &\leq \left(\frac{c(\alpha,\alpha')}{k+1}\right)^2 ||\mathbf{x}_k-\mathbf{x}^*||^2,
\end{split}
\end{equation}
where $c()$ is a constant which only depends on the parameter in the bracket. The proofs are given in \cite{Beck-AFastIterative} for the case $k'=0$. For the case when $k' \in \mathbb{N}>0$, the result can be  derived if we define a new sequence starting with $k'$.


\section{Weighted/Reweighted Compressed Sensing for Non-linear ECT}
\subsection{Weighted/Reweighted Compressed Sensing}
 The TV-IST algorithm presented in Section IV used the $l_1$ penalty on the total variation to achieve a better recovery. The recovery technique belongs to compressive sensing, which enables to reconstruct sparse signals exactly from what appear to be highly incomplete sets of linear measurements. Generally these problems can be solved by constrained $l_1$ minimization instead of the original $l_0$ penalty when the sensing scheme is appropriate. This is the theoretical foundation of the image reconstruction algorithm. However, in many cases $l_1$ minimization cannot achieve the exact sparse result that we want to pursue. The reason comes from the relationships between $l_0$, $l_1$ and $l_2$ norm. To understand why an $l_1$ but not $l_2$ minimization can achieve the same result of an $l_0$ minimization and the limitation of $l_1$, we can illustrate the optimization process as in Fig. \ref{fig:L0L1} and Fig. \ref{fig:wL1}, respectively.

In Fig. \ref{fig:L0L1}, the yellow lines represent the solutions for a given $l_2$ or $l_1$ norm, and the red lines are the solutions that satisfy the constraints. As a result, the meeting point of the yellow and red lines are the solution to this optimization problem. Comparing to the $l_2$ norm, the solution to $l_1$ norm is much more likely to lie on the axis, which implies it is a sparse solution.

In many complex problems dealing with high dimensional reconstruction, $l_1$ minimization also leads to a non-sparse solution. A simple 3-D example is illustrated in Fig. \ref{fig:wL1} \cite{Candes-Enhancing}, where $\mathbf{x}_0 = [0 \ 1 \ 0]^T$ and $\boldsymbol{\Phi} = \left[ \begin{array}{ccc} 2 \ 1 \ 1 \\ 1 \ 1\ 2 \end{array} \right]$. To recover $\mathbf{x}_0$ from $\mathbf{y}= \boldsymbol{\Phi} \mathbf{x}_0 = [1 \ 1]^T$, the real $\mathbf{x}_0$ is shown in (a). However, $l_1$ minimisation will give the wrong solution $\mathbf{x}' = [1/3 \ 0 \ 1/3]^T \neq \mathbf{x}_0$ when the interior of the $l_1$ ball intersects the feasible set $\boldsymbol{\Phi} \mathbf{x} = \mathbf{y}$ in (b), since $|\mathbf{x}'|_1 < |\mathbf{x}|_0$ and of course $\mathbf{x}'$ is not the sparse solution we need. In this case, instead of optimizing $ \min_\mathbf{x}\{\mathbf{y} - \boldsymbol{\Phi}\mathbf{x} + \alpha |\mathbf{x}|_1\} $, we can optimize the weighted $l_1$ norm as $ \min_\mathbf{x}\{\mathbf{y} - \boldsymbol{\Phi}\mathbf{x} + \alpha \sum_i \mathbf{w}^{(i)}|\mathbf{x}^{(i)}|\} $ where $\mathbf{w} \in \mathbb{R}^N$ is a weight vector over $\mathbf{x}$. If the weighting matrix $\mathbf{W} = \text{diag}(\mathbf{w}) = \text{diag}([3 \ 1 \ 3 ]^T)$, (c) shows the weighted $l_1$ ball of radius $|\mathbf{Wx}|_1 = 1$ centered at the origin and consequently we will find the correct solution $\mathbf{x}' = \mathbf{x}_0 $. People have shown that the same statements would hold true for any positive weighting matrix under certain conditions \cite{Candes-Enhancing}, and the weighted $l_1$ norm approach has been widely implemented \cite{Liu-AdaptiveWeightedTV,Guerquin-AFastWavelet, Candes-Enhancing}.

In order to solve the weighted $l_1$ problem, we modify the ISTA by adding a fixed weighting term $\mathbf{W}$ inside the $l_1$ norm and see $\mathbf{Wx}$ as the solution that needs to be calculated. Then the least squares term becomes $||\mathbf{y} - \boldsymbol{\Phi}\mathbf{W}^{-1} (\mathbf{Wx})||$. We follow ISTA to perform the updates. Yet here a reweighted algorithm will be proposed. The reweighted algorithm changes the weighting matrix $\mathbf{W}$ adaptively due to $\mathbf{x}$ to encourage few nonzero entries of $\mathbf{x}$.
In \cite{Candes-Enhancing}, a simple but effective iterative algorithm was proposed on which our algorithm is based. It alternately updates $\mathbf{x}$ and refines the weights $\mathbf{W}$. The algorithm consists of $2$ steps. \\

\begin{adjustwidth}{-0.1cm}{}
\begin{tabular}{l}
\hline
\textbf{Algorithm 3}: Iterative Reweighted Algorithm\\
\hline
\textbf{Input}: $\mathbf{y}, \boldsymbol{\Phi}$, $\rho$, \\
\textbf{Set}: max loop $k_{\text{max}}$. \\
\textbf{Initialize}: $\mathbf{w}_i =1, i = 1,\cdots, N$. \\
\textbf{Iteration}: for $k = 0,1, \cdots$, $k_\text{max}$ \ \ do \\
1. $\mathbf{x}_k = \arg\min_{\mathbf{x}}{||\mathbf{W}_k \mathbf{x}||_1} \ \text{s.t.} \ \mathbf{y} = \boldsymbol{\Phi} \mathbf{x}$ \\ $\backslash \backslash$ solve the weighted $l_1$ minimization  \\
2. $\mathbf{w}_k^{(i)} = \frac{1}{|\mathbf{x}_k^{(i)}|+ \rho}$ $\backslash \backslash$ update the weighted \\ for every $\mathbf{w}_k^{(i)}, i = 1,\cdots, N$  \\
\textbf{Output}: $\mathbf{x}$ . \\
\hline\\
\end{tabular}
\end{adjustwidth}

The parameter $\rho>0$ is introduced to provide stability and ensure a zero-valued component in $\mathbf{x}_k$ can also be modified as a nonzero estimate at the next step.

\begin{figure*}[hb]
\centering
\begin{tabular}{l}
\hline
\textbf{Algorithm 4}: Reweighted TV-FIST for non-linear ECT \\
\hline
\textbf{Input}: normalized sensitivity matrix $\mathbf{S}$, normalized capacitance $\boldsymbol{\lambda}$, \\ transform matrices $\mathbf{G}_1, \mathbf{G}_2$, permittivity range $\triangle \epsilon$, weight updating step $v$.\\
\textbf{Set}: max loop $k_{\text{max}}$, shrinkage parameter $\beta$, $\alpha'$, weighted parameter $\rho$. \\
\textbf{Initialize}: $\mathbf{g}_1, \mathbf{g}_2, \mathbf{h}_1, \mathbf{h}_2$ are zero vectors, $t_0 =1$,  $\mathbf{w} =\mathbf{1} \in \mathbb{R}^N$. \\
\textbf{Iteration}: for $k = 0,1, \cdots$, $k_\text{max}$ \ \ do \\
1. ${\mathbf{x}}_{k+1} = \mathbf{L}^{-1}(\mathbf{G}_1^T \mathbf{h}_1 + \mathbf{G}_2^T \mathbf{h}_2)$, $\backslash \backslash$ update the image according to auxiliary vectors \\
2. (\ref{eq:SS}) or (\ref{eq:S_ij}), $\backslash \backslash$ update sensitivity matrix to reduce non-linear effect \\
3. (\ref{eq:g12}) $\backslash \backslash$ calculate the steepest descent increment \\
4. $\hat{\mathbf{g}_1}_{k+1} = {\mathbf{h}_1}_{k} - \alpha'  \nabla {f_1}_1 $, $\hat{\mathbf{g}_2}_{k+1} = {\mathbf{h}_2}_{k} - \alpha' \nabla {f_1}_2 $, $\backslash \backslash$ \\
5. (\ref{eq:g=g1g2}), (\ref{eq:soft-th-g12-weight}),(\ref{eq:LS}) within $\triangle \epsilon$ threshold,(\ref{eq:gi})$\backslash \backslash$ repeat 3,4,5,6 steps in Algorithm 1 \\ using (\ref{eq:soft-th-g12-weight}) instead of (\ref{eq:soft-th-g12})\\
6. ${t}_{k+1} = \frac{1+\sqrt{1+4t_{k}^2}}{2}$,  $ \backslash \backslash$  calculate $t$\\
7. ${\mathbf{h}_i}_{k+1} = {\mathbf{g}_i}{k+1} + \left( \frac{t_{k}-1}{t_{k+1}} \right)({\mathbf{g}_i}_{k+1}- {\mathbf{g}_i}_{k}), i \in \{1,2\}$, $ \backslash \backslash$  calculate $\mathbf{h}_1, \mathbf{h}_2$ \\
8. Every $v$ iteration, $\mathbf{w}_k^{(i)} = \frac{1}{|\mathbf{x}_k^{(i)}|+ \rho}$ $\backslash \backslash$ update the weights \\
\textbf{Output}: $\mathbf{x}$ or $\mathbf{X}$. \\
\hline\\
\end{tabular}
\end{figure*}

\subsection{Reweighted TV-IST for ECT}
Similar to the weighted $l_1$ minimization, it is natural to incorporate the reweighting technique in the total variation constraints. Then the TV optimization problem for ECT is transformed to
\begin{equation}\label{eq:TVIST2}
\mathbf{x} = \arg \min_{\mathbf{x}}{ ||\boldsymbol{\lambda} - \mathbf{S}\mathbf{L}^{-1}(\mathbf{G}^T_1 \mathbf{g}_1 + \mathbf{G}_2^T \mathbf{g}_2)||^2 + \alpha||\mathbf{Wg}||_{1}}.
\end{equation}
 In the following steps, (\ref{eq:g12}) and (\ref{eq:hat_g12}) remain the same for reweighted TV-IST in each iteration. The difference occurs in the soft thresholding step. Because we want to pursue the minimal $l_1$ norm of the weighted TV, the thresholding process needs to be changed adaptively:
\begin{equation}\label{eq:soft-th-g12-weight}
\begin{split}
{\mathbf{g}_1}^{(i)}_{k+1} &= \frac{\mathcal{T}_{\alpha'}(\mathbf{W}\hat{\mathbf{g}}^{(i)}_{k+1})}{\mathbf{Wg}^{(i)}_{k+1}} \hat{\mathbf{g}_1}^{(i)}_{k+1}  \\
{\mathbf{g}_2}^{(i)}_{k+1} &= \frac{\mathcal{T}_{\alpha'}(\mathbf{W}\hat{\mathbf{g}}^{(i)}_{k+1})}{\mathbf{Wg}^{(i)}_{k+1}} \hat{\mathbf{g}_2}^{(i)}_{k+1}.
\end{split}
\end{equation}
where this operation should be done element-wise. It is a weighted version of the 2D soft threshold update. By using the weighted $\mathbf{Wg}$ for thresholding, the $l_1$ norm behaves more like the $l_0$ norm. All the non-zero entries of $\mathbf{g}$ above the threshold will be calculated more equally in the weighted norm, similar to the definition of the $l_0$ norm which see all non-zero entries contribute equally. Finally, the weights $\mathbf{w}$ vary depending on $\mathbf{g}$. Specifically, the weights can be updated element by element as the second step of Algorithm 3:
 \begin{equation}
 \mathbf{w}_{k+1}^{(i)} = \frac{1}{|\mathbf{g}_k^{(i)}|+ \rho},
 \end{equation}
 where $\rho>0$ is a parameter that is set slightly smaller than the expected nonzero magnitudes of $\mathbf{g}$. The value for $\rho$ can be determined from experience, but in general should be small. 
 Moreover, in practice since the reconstructed result $\mathbf{x}$ and its gradient $\mathbf{g}$ evolves gradually after each iteration, we insert the weights updating step into the TV-IST algorithm every $v$ iterations. Hence the weights can be updated every $v$ iterations and we can use the parameter $v$ to make a tradeoff between calculation speed and weights update. Meanwhile, the auxiliary vectors $\mathbf{h}_1, \mathbf{h}_2$ can also be adopted to accelerate the convergence and the reweighted TV-IST becomes reweighted TV-FIST. The faster implementation is always used here and henceforth the reweighted TV-FIST will be referred to as reweighted TV-IST for simplicity.

\subsection{Reweighted TV-IST Algorithm for Non-linear ECT}
Before demonstrating the reweighted TV-IST algorithm, we explain two techniques that can be used to compensate for the non-linearity in our algorithm. In the ECT model, the non-linear effects can be resolved  in two aspects. The first one is the approximation of the linear model in (\ref{eq:linearC}). The second non-linear effect lies on the accuracy of the measurements of the sensitivity matrix $\mathbf{S}$. Two techniques are introduced here to address the non-linear effect, respectively \cite{Villares-ANonLinear,Li-ImageRecon}. However, only the first of these is implemented in the simulations as it is computationally simpler.

\begin{figure*}[ht]
   \centering
   \includegraphics[width=12.8cm]{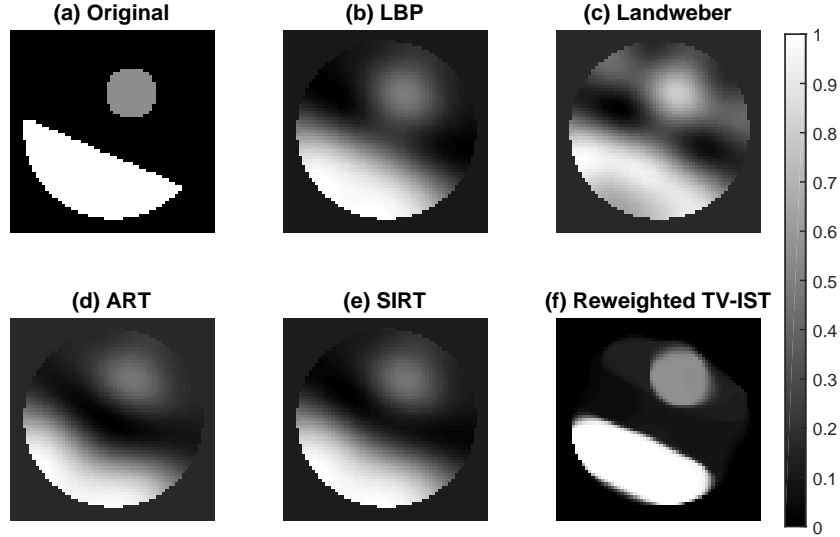} 
  \caption{Simulation illustrating ECT reconstruction of the `two bubbles' phantom. (a) The original phantom image. The reconstructions are based on the capacitance measurements with Gaussian noise added , the signal-to-noise ratio is 35 dB. Reconstructions of the phantom are shown for (b) LBP reconstruction, (c) Landweber (d) ART, (e) SIRT, (f) Reweighted TV-IST.}\label{fig:1}
\end{figure*}

From (\ref{eq: triangleC}) to (\ref{eq:linearC}) the quadratic and higher order terms have been neglected to reduce the ECT model to a linear model. However this approximation causes errors due to the higher order terms that have been neglected. To offset this bias, a fitting curve has been proposed \cite{Villares-ANonLinear}:
\begin{equation}\label{eq:m}
\triangle \mathbf{m}_l \approx \frac{\triangle {\epsilon} \cdot 2 {\epsilon} \mathbf{S}_l V}{ \triangle {\epsilon} +2{\epsilon}},  \ \ l = 1,\cdots , N(N-1)/2,
\end{equation}
where $\mathbf{m}_l$ denotes the measurement at electrode $i$ when electrode $j$ is under the voltage $V$ while other electrodes are grounded, $l = (i-1)N+j, i<j$. This setting is to make sure that $\triangle \mathbf{m}_l = 2 \epsilon \mathbf{S}_l V$ when $\triangle \epsilon$ tends to infinity and the slope is $\mathbf{S}_l V$ at $\triangle \epsilon \rightarrow 0$. This approach may reduce the non-linear error by around $10\%$ \cite{Villares-ANonLinear}. The non-linear sensitivity matrix $\hat{\mathbf{S}}$ can be defined as
\begin{equation}\label{eq:S}
\hat{\mathbf{S}}_l = \frac{\triangle \mathbf{m}_l } {V \triangle \epsilon}.
\end{equation}
Combining (\ref{eq:m}) and (\ref{eq:S}) we have
\begin{equation}\label{eq:SS}
\hat{\mathbf{S}}_l = \frac{2\epsilon}{\triangle \epsilon + 2 \epsilon} \mathbf{S}_l,
\end{equation}
which adjusts $\mathbf{S}$ to a non-linear sensitivity matrix $\hat{\mathbf{S}}$, where the permittivity of the area of interest is assumed to vary from $\epsilon$ to $\triangle \epsilon$, whose values can be determined before the experiments.

An adaptive sensitivity matrix model has also been proposed for use with Landweber iterations \cite{Li-ImageRecon}. We introduce this feedback iteration to our reweighted TV-IST algorithm. The sensitivity map for an electrode pair can be calculated from the potential distribution
\begin{equation}\label{eq:S_ij}
\mathbf{S}_{i,j}(r,c) = - \oint_{(r,c)} \left( \frac{\partial \phi_i}{ \partial r} \cdot \frac{\partial \phi_j}{ \partial r} + \frac{\partial \phi_i}{ \partial c} \cdot \frac{\partial \phi_j}{ \partial c} \right) \text{d}r \text{d}c,
\end{equation}
where $\frac{\partial \phi_i}{ \partial r}, \frac{\partial \phi_i}{ \partial c}$ are the gradient values of the potential with electrode $i$ in the row and column vectors, respectively; and the potential value of each pixel can be computed after iterations using the finite difference method (FDM) depending on the potential values of the surrounding four pixels:
\begin{equation}
\begin{split}
P_1 &= \phi_{i-1,j} \epsilon_{i-1,j} ; \ \  P_2 = \phi_{i+1,j} \epsilon_{i+1,j} ;  \\
P_3 &= \phi_{i,j-1} \epsilon_{i,j-1} ; \ \  P_4 = \phi_{i,j+1} \epsilon_{i,j+1};\\
\phi_{i,j} &= \frac{P_1 + P_2 + P_3 +P_4  }{\epsilon_{i-1,j}+\epsilon_{i+1,j}+\epsilon_{i,j-1} +\epsilon_{i,j+1}}.
\end{split}
\end{equation}
where $i,j$ are the location indexes.

In summary, the reweighted TV-IST algorithm for non-linear ECT uses the TV penalties in the cost function to pursue the optimal solution iteratively and meanwhile makes use of the superiority of updated reweighted norms and the auxiliary method's fast convergence. It is distinct from conventional FISTA for total variation minimization and designed to be suitable for ECT reconstruction specifically. Compared to the conventional linear TV-IST, our non-linear reweighted TV-IST has two differences. Firstly the reweighted term $\mathbf{w}$ has been adopted in the cost function to pursue a more sparse total variation in the optimization process. This should produce clearer edges between areas with different permittivities. Secondly, two methods are introduced to reduce the non-linear effects. The methods add an extra step in the Algorithm (step 2 in Algorithm 4) to update the sensitivity matrix during the calculation. The two methods of non-linearity correction introduced in this section have similar effects. One is derived from the second order terms of the Taylor series expansion, while the other represents the non-linearity from a potential distribution perspective. In Algorithm 4, either (\ref{eq:SS}) or (\ref{eq:S_ij}) can be used to compensate the non-linearity in (\ref{eq:C}). Herein we only consider the correction obtained using (\ref{eq:SS}) as this implementation is faster.

\begin{figure*}[ht]
   \includegraphics[width=15.5cm,left]{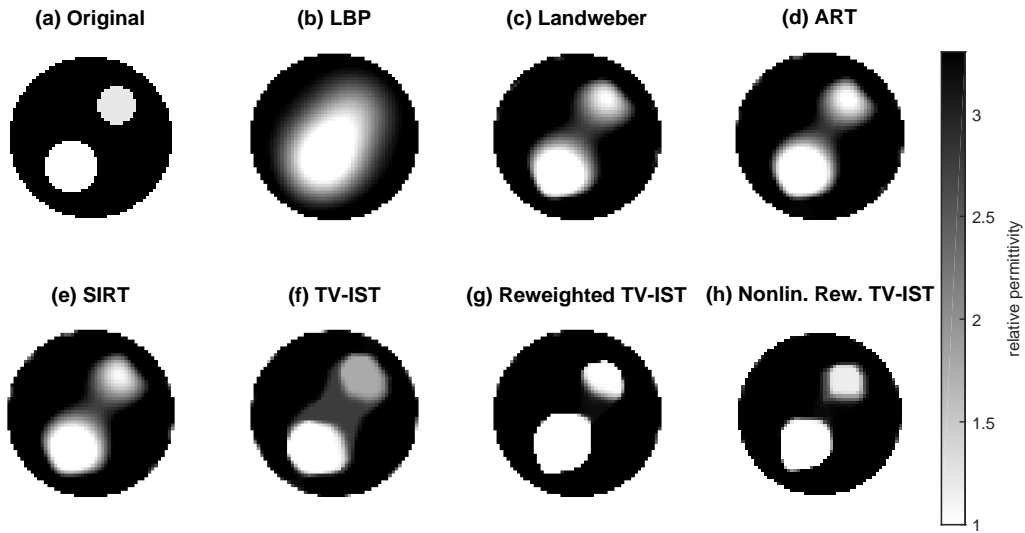} 
  \caption{Simulation illustrating ECT reconstruction of the 'two bubbles' phantom. (a) The original phantom image. Reconstructions of the phantom are shown for (b) LBP reconstruction, (c) Landweber (d) ART, (e) SIRT, (f) TV-IST, (g) Reweighted TV-IST, (h) Reweighted TV-IST for Non-linear ECT using (\ref{eq:SS}). Note that the colour scale is inverted compared with Fig. \ref{fig:1}. }\label{fig:WwoBubble}
\end{figure*}

\section{Experiments}
To test the proposed algorithm, numerical simulations were performed on an ECT model. The results of reweighted TV-FIST algorithm are compared with the performances of several widely used algorithms in practice, which include LBP, ART (relaxed Kacmarz iteration) and SIRT (relaxed Cimmino iteration). All reconstructions are carried out on a standard desktop PC with an AMD Phenom(tm) $3.0$ GHz processor and $7.8$ GB RAM. The simulations are run in MATLAB 2009b and $500$ iterations were run throughout for each of these algorithms. The ECT system (normally with $8$ or $12$ electrodes here) was modeled using the Comsol Multiphysics software package, and the sensitivity matrix was generated from Comsol for all reconstructions. As in \cite{Chandrasekera-MeaOfBubble}, the normalised capacitance was used to help minimise the effect of non-linearity introduced by the wall of the sensor \cite{Yang-AnImageRec}.

Firstly, we implement various algorithms on a phantom consisting of an arc-shaped part and a circular object in a $64 \times 64$ pixel image, as shown in Fig. \ref{fig:1}. In the ECT system we use $8$ electrodes, which can provide $28$ independent capacitance measurements. The smaller round object has image intensity (a.u.) of $0.6$ and the larger object has an intensity of $1$; black and white in Fig. \ref{fig:1}  correspond to intensities of 0 and 1, respectively. From Fig. \ref{fig:1} (b) one can see that the two objects can be recovered approximately by the LBP reconstruction, however the shape is significantly smoothed and broadened compared with the true image in (a). The Landweber, ART and SIRT methods show a similar recovered result in (c-e). Errors in the permittivity distribution make precise identification of the boundary of the objects challenging. Fig \ref{fig:1} (f) shows the reconstruction using reweighted TV-IST. The boundaries of both objects are clearly resolved with the correct intensity. The only significant error occurs at the wall of the system. The error at the wall is likely caused by non-linearity at the wall, or the use of the isotropic form of TV which can introduce smoothing at sharp points in the image.

In the second simulation, the ECT system consists of $12$ electrodes, which can provide $66$ independent capacitance measurements. The tested permittivity distribution was the `two bubbles' image, as shown in Fig. \ref{fig:WwoBubble}(a). It is a phantom image consisting of a circular pipe containing two circular objects in a $64 \times 64$ pixel image.
In the simulation, the relative permittivity of the cylindrical wall and the background was set to $3.3$; the relative permittivity of the two circular objects were {$1$} and {$1.22$} for the large and small objects, respectively.
Fig. \ref{fig:WwoBubble}(a) is different from the normal ECT permittivity distributions considered since the background has a high permittivity while the two bubbles have low permittivity. The LBP reconstruction of this image, shown in Fig. \ref{fig:WwoBubble}(b), fails as both bubbles blur into a single object. The poor reconstruction arises from the close proximity of the two bubbles and the use of a high permittivity background.  Landweber iterations, Fig. \ref{fig:WwoBubble}(c), gives a better result with the two bubbles resolved, but the bubbles still appear heavily smoothed. Similar results were obtained for the ART and SIRT reconstructions. The linear TV-IST algorithm, Fig. \ref{fig:WwoBubble}(d), recovers the sharp boundaries around the two bubbles. However, a high permittivity ``bridge'' is seen connecting the two bubbles and the permittivity of the smaller bubble is over estimated. The proposed reweighted TV-IST result is shown in Fig. \ref{fig:WwoBubble}(g). The outline of the two bubbles is recovered fairly accurately, with only a slight tendency of the two bubbles to merge together and the size of the two bubbles overestimated by $\sim 11$\%.  The ``bridge'' seen using the standard TV-IST algorithm has been eliminated. The permittivity is also recovered fairly well with the permittivity of the large bubble found to be $1$ and the permittivity of the small bubble $1.05$, which compare with the input permittivities of $1$ and $1.22$, respectively. The non-linear reweighted reconstruction is shown in Fig. \ref{fig:WwoBubble}(h). The recovered bubble shapes are slightly more ``square'' than the input bubble shapes, but otherwise the outline of both bubbles is recovered well. The size of each bubble is accurate to within $5$\% of the true bubble size.  The permittivity in the large and small bubbles was $1$ and $1.17$, respectively, in good agreement with the true values. The reconstruction quality is sensitive to the choice of the parameters $\beta$, $\alpha'$, and $\rho$, as well as the number of iterations performed. However, overall these results demonstrate that the introduction of the reweighted TV-IST algorithm, including non-linearity correction, significantly improves the quality of the reconstructed images for piecewise smooth input permittivity distributions. The re-weighting approach enables the solution to approach the true $l_0$-norm solution closely, while the updates to the sensitivity matrix during image reconstruction help mitigate against the non-linearity effects.

\section{Conclusion}
In this paper, a non-linear reweighted total variation algorithm for reconstruction of images obtained from ECT measurements has been proposed and analyzed. The proposed algorithm penalises the $l_1$-norm of the spatial finite differences of the image (total variation) by using an iterative thresholding approach. A varying weight calculated in each iteration is used to make sure that the result converges towards the desired $l_0$-norm. In addition, the non-linearity of the governing equations was considered and a straightforward approach to update the sensitivity matrix was introduced accordingly. The proposed algorithm was verified on two simulated permittivity distributions. It is shown that the reweighting significantly increases the quality of the reconstructed images recovering sharper boundaries with fewer artefacts than existing algorithms including LBP, ART, SIRT and our previous implementation of TV-IST. The incorporation of the updated sensitivity matrix to approximate the non-linearity of the ECT sensor further increased the accuracy of the reconstructed images, most notably in recovering quantitative permittivity values in each domain. The new algorithm here promises to increase the quality of ECT imaging. We anticipate even greater benefits if the algorithm can be combined with recently proposed enhanced sensing strategies~\cite{Fan-EnhancementOf, Marashdeh-AECVT}.

\section*{Acknowledgement}
This work was partially supported by the EPSRC Grant Reference: EP/K008218/1. The authors would like to thank T.C. Chandrasekera and Yi Li for assisting with the comparison to existing image reconstruction algorithms.

\bibliographystyle{IEEEtran}

\end{document}